\documentclass[10pt,twocolumn,letterpaper]{article}

\usepackage[pagenumbers]{cvpr} 
\usepackage{pifont}

\usepackage{multirow}
\usepackage{array}
\usepackage{algorithm}
\usepackage{algorithmicx}
\usepackage{algpseudocode}
\usepackage[dvipsnames]{xcolor}

\definecolor{cvprblue}{rgb}{0.21,0.49,0.74}
\usepackage[pagebackref,breaklinks,colorlinks,citecolor=cvprblue]{hyperref}

\def\paperID{3070} 
\def\confName{CVPR}
\def\confYear{2024}

\title{ADM-Loc: Actionness Distribution Modeling \\ for Point-supervised Temporal Action Localization}

\author{Elahe~Vahdani \quad Yingli~Tian \\
The City University of New York \\
}

\begin{document}
\maketitle


\begin{abstract}
This paper addresses the challenge of point-supervised temporal action detection, in which only one frame per action instance is annotated in the training set. Self-training aims to provide supplementary supervision for the training process by generating pseudo-labels (action proposals) from a base model. However, most current methods generate action proposals by applying manually designed thresholds to action classification probabilities and treating adjacent snippets as independent entities. As a result, these methods struggle to generate complete action proposals, exhibit sensitivity to fluctuations in action classification scores, and generate redundant and overlapping action proposals. This paper proposes a novel framework termed ADM-Loc, which stands for \textbf{A}ctionness \textbf{D}istribution \textbf{M}odeling for point-supervised action \textbf{Loc}alization. ADM-Loc generates action proposals by fitting a composite distribution, comprising both Gaussian and uniform distributions, to the action classification signals. This fitting process is tailored to each action class present in the video and is applied separately for each action instance, ensuring the distinctiveness of their distributions. ADM-Loc significantly enhances the alignment between the generated action proposals and ground-truth action instances and offers high-quality pseudo-labels for self-training. Moreover, to model action boundary snippets, it enforces consistency in action classification scores during training by employing Gaussian kernels, supervised with the proposed loss functions. ADM-Loc outperforms the state-of-the-art point-supervised methods on THUMOS’14 and ActivityNet-v1.2 datasets.
\end{abstract}


\section{Introduction}


Automated video analysis has broad applications in computer vision research, benefiting diverse fields like self-driving cars, public safety monitoring, and sports analysis\cite{he2018anomaly, cioppa2020context, giancola2018soccernet, rasouli2019autonomous, mahadevan2019av, yao2020and}. A principal challenge in this field is Temporal Action Localization (TAL) in untrimmed video streams, with the objective being to accurately pinpoint the start and end times of actions and to categorize them accordingly \cite{vahdani2023deep, wang2023temporal}. Recent advancements in fully-supervised TAL methods have shown promising improvements \cite{zhang2022actionformer,wang2023videomae, shi2023tridet}. However, they depend on the detailed annotation of start and end timestamps, along with action labels for every action in training videos, which is both labor-intensive and expensive. To diminish the dependence on extensive labeling throughout the training stage, there has been a growing interest in the advancement of methodologies that operate under limited supervision \cite{rizve2023pivotal,zhou2023improving, ren2023proposal,chen2022dual}. Specifically, point-supervised TAL requires the annotation of only a single frame within the temporal window of each action instance in the input video \cite{moltisanti2019action, ma2020sf, lee2021learning, yang2021background, li2023sub}. Point-level supervision significantly lowers the annotation costs in comparison to full supervision, while providing essential information about the approximate locations and the total number of action instances.

In temporal action detection, pseudo-labels are primarily defined as estimated action boundaries (proposals) along with their corresponding action labels. A recent trend aimed at bridging the gap between point-supervised and fully-supervised TAL relies on self-training, wherein pseudo-labels are generated by a base point-supervised model. These pseudo-labels act as substitute action annotations, enabling the training of models under limited supervision. Current techniques generate pseudo-labels by creating proposals based on thresholds applied to the predicted action classification probabilities. However, these methods have several shortcomings. Firstly, they are highly sensitive to the choice of threshold values; varying thresholds can lead to significant shifts in the alignment of proposals with ground-truth instances. Secondly, they often yield an excess of redundant and overlapping proposals, which are unsuitable as pseudo-labels. Ideally, there should be a one-to-one correspondence between pseudo-labels and action instances. Lastly, these methods struggle to generate complete action proposals and are sensitive to inconsistencies in action classification scores.


We introduce an innovative approach to generate pseudo-labels by modeling the distribution of action classification probabilities as a combination of Gaussian and uniform distributions. This methodology is based on the observation that certain action instances exhibit homogenous classification probabilities across snippets, resembling a uniform distribution. In contrast, for other actions, snippets near the action boundaries, which often include ambiguous or transitional movements, show lower classification probabilities, resembling a Gaussian distribution. This combination effectively captures the full spectrum of action instances. Our base point-supervised model predicts background snippets and action classification probabilities for each action class in the video. For each annotated action point, preliminary action boundaries are determined by identifying the nearest background timestamps before and after the annotated point. Then, a mixed distribution model is fitted to the action classification probabilities within these boundaries, minimizing the mean squared error (MSE) loss using Brent's method \cite{brent2013algorithms}. Consequently, high-quality pseudo-labels are generated that overcome prior challenges: 1) eliminating reliance on arbitrary thresholding, 2) ensuring the creation of a single proposal for each action instance, and 3) maintaining robustness against fluctuations in action classification probabilities. Additionally, we propose learning action boundary snippets during the training of the main model by modeling the distribution of action scores. Although snippets near the action boundaries often have lower classification scores compared to more central action snippets, differentiating these boundary snippets from the background is essential. During training, we compare the predicted classification probabilities with the Gaussian kernels to reinforce the consistency of action scores across the entire range of actions, including boundaries. This process, supervised with our proposed loss functions, enhances the model's accuracy in estimating action durations and in generating complete proposals. Our contributions are summarized as follows:

\begin{itemize}
    \item We propose a novel strategy for pseudo-label generation in self-training, where the predicted action classification probabilities are modeled as a composite of Gaussian and uniform distributions. The effectiveness of the strategy is evidenced by the high-quality pseudo-labels it generates.
    \item We propose a framework of learning action boundary snippets during the training of the main model to generate complete action proposals for testing. This process involves comparing the predicted action classification probabilities with a Gaussian kernel predicted by our model. Our designed loss functions supervise the learning of Gaussian parameters and the predicted probability signals.
    \item Our ADM-Loc framework outperforms the state-of-the-art point-supervised methods on THUMOS'14 and ActivityNet-v1.2 datasets. 
\end{itemize}

\section{Related Work}

\textbf{Fully-supervised TAL.} Fully-supervised methods can be grouped into anchor-based and anchor-free. Anchor-based methods generate pre-defined action proposals distributed across temporal locations \cite{gao2017turn,gao2017cascaded,chao2018rethinking}. They extract fixed-size features from the proposals to evaluate their quality. Anchor-free methods generate proposals with flexible duration by predicting actionness and action offset for each snippet. \cite{lin2018bsn, lin2019bmn, lin2020fast,lin2021learning, bai2020boundary}. Temporal feature pyramid is introduced to model actions of varying duration \cite{lin2017single, zhang2018s3d,liu2019multi, liu2020progressive}. Modeling temporal dependencies in videos has been addressed by recurrent neural networks \cite{buch2017sst, buch2019end}, graph convolutions \cite{zeng2019graph, li2020graph, bai2020boundary, xu2020g, zhao2020video}, and transformers \cite{zhang2022actionformer, nawhal2021activity,chang2022augmented}. Unlike these methods that require detailed frame-level annotations, our framework relies solely on point-level annotations. We employ a multi-scale transformer architecture to model the temporal dependencies of video snippets and to handle actions of varying durations.


\textbf{Weakly-supervised TAL.} The methods often require only the video-level labels of actions for training, while the temporal boundaries of actions are not needed. Majority of the weakly-supervised methods rely on the Multi-Instance Learning (MIL) to learn actionness and classification scores to detect discriminative action regions and eliminate background snippets \cite{lee2020background, narayan20193c, hong2021cross, luo2021action, narayan2021d2, qu2021acm}. To generate complete action proposals, some methods have proposed adversarial complementary learning approaches to discover different parts of actions by increasing the weight of less
discriminative parts of the video \cite{singh2017hide,zhong2018step, min2020adversarial,zeng2019breaking, liu2019completeness}. Another category of methods rely on self-training scheme to generate pseudo-labels on the train set from an initial base model. The pseudo-labels provide additional supervision for the main model to improve the training \cite{he2022asm,luo2020weakly,pardo2021refineloc, yang2021uncertainty,zhai2020two, rizve2023pivotal}. These methods often fail to generate high-quality pseudo-labels. In contrast, our model, employing slightly more annotations, produces pseudo-labels that are significantly better aligned with the ground-truth action instances.

\begin{figure*}[t]
\begin{center}
   \includegraphics[width=\textwidth]{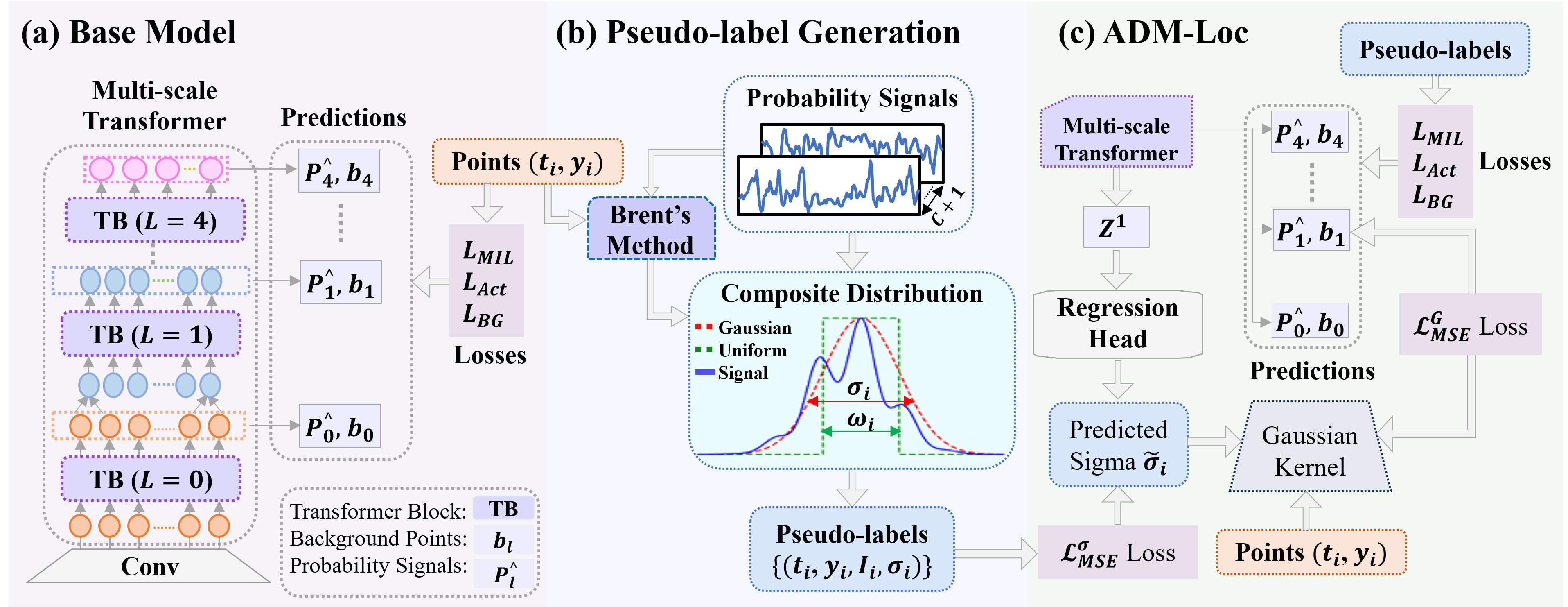}
\end{center}
   \caption{ Framework overview. (a) The base model is a multi-scale transformer supervised with point-level labels and $\mathcal{L}_{\text{MIL}}$, $\mathcal{L}_{\text{Act}}$, and $\mathcal{L}_{\text{BG}}$ losses. (b) The  probability signals predicted by the base model are given to our Actionness Distribution Modeling (ADM) for pseudo-label generation. A composite distribution, comprising both Gaussian and uniform distributions, is fitted to the predicted probability signals and optimized by Brent's method \cite{brent2013algorithms} for pseudo-label generation. (c) ADM-Loc is a multi-scale transformer supervised with our generated pseudo-labels and optimized by $\mathcal{L}_{\text{MIL}}$, $\mathcal{L}_{\text{Act}}$, $\mathcal{L}_{\text{BG}}$, $\mathcal{L}^{\sigma}_{\text{MSE}}$ and $\mathcal{L}^{G}_{\text{MSE}}$ losses. ADM-Loc learns action boundary snippets by comparing the predicted action classification probabilities with a predicted Gaussian kernel, supervised by our proposed loss functions.}
\label{fig:framework}
\end{figure*}

\textbf{Point-supervised TAL.} Point-level supervision significantly reduces the cost of annotation by labeling a single point for each action instance. SF-Net \cite{ma2020sf} proposed to expanded each annotated single frame to its nearby frames to mine pseudo action frames and utilized the unannotated frames to mine pseudo background frames. PTAL \cite{ju2020point} performed boundary regression based on keyframe prediction. Back-TAL \cite{yang2021background} introduced background-click supervision by annotating a random frame from a series of consecutive background frames. Lee \textit{et al.} \cite{lee2021learning} developed an action-background contrast method to capture action completeness. We propose a novel approach for generating high-quality pseudo-labels using a base point-supervised model. These pseudo-labels then guide our main model in learning action continuity and in generating complete action proposals during testing.


\section{Our Proposed Method}

Fig.~\ref{fig:framework} provides an overview of our framework. Our framework adopts a self-training strategy that incorporates a base model and a main model, each employing a multi-scale transformer as their backbone architecture. The base model's objective is to predict action probability signals and background points, Fig.~\ref{fig:framework}(a). The predicted probability signals are employed to generate high-quality pseudo-labels, providing additional supervision for the training of the main model (ADM-Loc), Fig.~\ref{fig:framework}(b). ADM-Loc learns action boundary snippets by comparing the predicted probabilities with a predicted Gaussian kernel supervised by our proposed loss functions, $\mathcal{L}^{\sigma}_{\text{MSE}}$ and $\mathcal{L}^{G}_{\text{MSE}}$, Fig.~\ref{fig:framework}(c). 

\subsection{Point-Supervised Formulation} 

Given an input video, a single annotated point with the action category is provided for each action instance, denoted by \(\{t_i, y_i\}_{i=1}^{N_{\text{act}}}\). The \(i\)-th action instance is annotated at the \(t_i\)-th snippet with its action label \(y_i\), and \(N_{\text{act}}\) is the total number of action instances in the input video. The label $y_i$ is a binary vector with $y_i[c] = 1$ if the $i$-th action instance belongs to class $c$ and otherwise 0 for $C$ action classes.

\subsection{Backbone Architecture}
\label{Backbone}

A multi-scale temporal transformer is employed as the backbone architecture. Given an input video, snippet-level visual features are extracted with a pre-trained visual encoder (I3D \cite{carreira2017quo}) and concatenated to generate a video feature sequence $X \in \mathbb{R}^{T \times D}$, where $T$ is the number of snippets and $D$ is the feature dimensionality. Each snippet feature is embedded using a shallow temporal convolutional network resulting in feature sequence $Z^0 \in \mathbb{R}^{T \times D}$. This feature sequence is the input to the transformer network to model the temporal dependencies using local self-attention \cite{zhang2022actionformer}. To represent actions with different duration, a temporal feature pyramid is constructed by down-sampling transformer blocks using a strided depthwise 1D convolution. The feature pyramid is denoted by $Z =\{Z^1,Z^2,\cdots,Z^L\}$ where $Z^l \in \mathbb{R}^{T_l \times D}$ is the output of level $l$. Also, $T_l = T/{\theta^l}$, and $\theta$ is the down-sampling ratio. Feature pyramid captures multi-scale temporal information, enabling the model to capture both short-term and long-term temporal dependencies, leading to a more comprehensive representation of action dynamics. A shallow 1D convolutional network is attached to each pyramid level with its parameters shared across all levels. A sigmoid function is attached to each output dimension to predict the probability of actions and background. The output of the $l$-th level of the feature pyramid is a probability sequence, denoted by $P_l \in \mathbb{R}^{ T_{l}\times {C+1}}$, where $T_l$ is the temporal dimension on the $l$-th level. Additionally, $P_l[t,C+1]$ is the probability of background at time $t$ on level $l$. The complement of the background probability is the class-agnostic score. The class-specific and class-agnostic scores are fused to derive the final probability sequence $\hat{P_l} \in \mathbb{R}^{T_l \times C+1}$.
\vspace{-3pt}
\begin{equation}
    \hat{P_l}[t,c] = P_l[t,c] (1-P_l[t,C+1]).
\end{equation}

\subsection{Point-supervised Base Model}
\label{base-mode}

\textbf{Augmented annotations}. We augment the point-level annotations for improved training by defining a vicinity around each annotated point with a hyper-parameter radius \( r_a \). Specifically, for the \( i \)-th action instance containing the annotated point \( t_i \) and its corresponding label \( y_i \), the label \( y_i \) is assigned to all snippets within radius $r_a$. The augmented annotation set is denoted by $\Phi$. 

\begin{equation}
    \Phi = \{([t_i-r_a, t_i+r_a], y_i)\}_{i=1}^{N_\text{act}}. 
\end{equation}


The augmented annotation set on level $l$ is defined as the following where $\theta$ represents the down-sampling ratio. The notation is simplified on the second line. $N_l$ is the number of labeled points on level $l$ after augmentation.

\begin{equation}
\begin{aligned}
    \Phi^l & =  \{([(t_i/\theta^l)-r_a, (t_i/\theta^l)+r_a], y_i)\}_{i=1}^{N_\text{act}} \\
    & =  \{(t_j, y_j)\}_{j=1}^{N_l}. 
\end{aligned}
\end{equation}


\textbf{Video-level action prediction}. The video-level score for class $c$ is defined as the average of action probabilities for class c over the top-$k$ temporal positions on each level $l$ of the pyramid, denoted by $\Bar{P}_l[c]$. The Multiple Instance Learning (MIL) loss \cite{dietterich1997solving} is utilized to supervise the predictions. The video-level label is denoted by $y$.

\vspace{-5pt}
\begin{equation}
  \mathcal{L}_{\text{MIL}} = - \frac{1}{L} \sum_{l=1}^{L} \sum_{c=1}^{C} y[c] \log(\Bar{P}_l[c]) + (1-y[c]) \log (1-\Bar{P}_l[c]).
\label{eq.MIL}
\end{equation}

\textbf{Snippet-level action prediction}. The snippet-level focal loss is employed to optimize the probability signal $\hat{P_l}$ for each level $l$ of the pyramid. $\gamma$ is the focusing parameter (set to 2) and $N^{\star}_{\text{act}}$ is the number of positive instances.

\begin{equation}
\resizebox{0.95\linewidth}{!}{$
\begin{aligned}
\mathcal{L}_{\text{Act}} = -\frac{1}{N^{\star}_{\text{act}}} &\sum_{l=1}^{L} \sum_{j=1}^{N_l} \sum_{c=1}^{C} y_j[c] \log(\hat{P}_l[t_j,c])  (1-\hat{P}_l[t_j,c])^{\gamma}\\
&- (1-y_j[c]) \log (1-\hat{P}_l[t_j,c]) \hat{P}_l[t_j,c]^{\gamma}.
\end{aligned}
$}
\label{act-loss}
\end{equation}

\textbf{Background prediction}. To distinguish actions from the background, we select the temporal positions not belonging to any of the augmented annotated points and possessing a background probability exceeding a certain threshold on each level $l$ of the pyramid. The background points on level $l$ are denoted by $\{b_j\}_{j=1}^{M_l}$ with $p_l(b_j)$ as the probability of background at time $b_j$. The background loss is employed to optimize the probability signals $\hat{P_l}$ for all levels. $M_{\text{bg}}$ is the total number of background points. 

\begin{align}
  \footnotesize
    \mathcal{L}_{\text{BG}} = & - \frac{1}{M_{\text{bg}}} \sum_{l=1}^{L}  \sum_{j=1}^{M_l} \Bigl[ 
    \sum_{c=1}^{C} (\hat{P}_l[b_j,c])^{\gamma} \log (1-\hat{P}_l[b_j,c]) \nonumber \\
    & + (1-p_l(b_j))^{\gamma} \log p_l(b_j).
    \normalsize
\label{bg-loss}
\end{align}

\textbf{Joint training}. The total loss for the base model is a weighted combination of the three aforementioned losses where $\lambda_{\star}$ terms are determined through empirical analysis.

\begin{align}
  L_{\text{Total}} = \lambda_{\text{MIL}}\mathcal{L}_{\text{MIL}} + \lambda_{\text{Act}} \mathcal{L}_{\text{Act}} + \lambda_{\text{BG}} \mathcal{L}_{\text{BG}}.  
  \label{total-loss}
\end{align}


\subsection{Actionness Distribution Modeling (ADM)}
\label{ADM-section}

\subsubsection{Pseudo-label Generation with ADM}
\label{PLG-ADM-section}

Our proposed pseudo-labels generation method on the training set models the distribution of action classification probabilities predicted by the base model. This distribution is represented as a combination of Gaussian and uniform distributions. The rationale behind this modeling is that certain action instances exhibit uniform classification probabilities across snippets, resembling a uniform distribution. Conversely, actions with ambiguous boundaries or transitional movements tend to have lower classification probabilities near the boundaries, indicative of a Gaussian distribution. This combination of distributions captures the full spectrum of action instances. 

After training the base model, action classification probabilities are extracted from the final level of the multi-scale transformer, denoted by $\hat{P_L} \in \mathbb{R}^{T_L \times C+1}$. This choice is made because the larger receptive field at the last feature pyramid level exhibits fewer fluctuations in action probabilities across neighboring snippets, making it more suitable for our modeling purposes. A Gaussian filter is also applied to smooth the signal and reduce the impact of minor inconsistencies in action classification probabilities. The resolution of the last-level probability signal is upgraded to match that of the first level, resulting in signal \( \tilde{P}_L \in \mathbb{R}^{T \times (C+1)} \). 

The background points are predicted from the first level of the pyramid because the lower resolution of the first level excels at detecting fine-grained information. The annotated action points and the predicted background points are denoted by \( \{ (t_i, y_i) \}_{i=1}^{N_{\text{act}}} \), and \( \{ b_j \}_{j=1}^{N_{\text{bkg}}} \), respectively. For each annotated action point $(t_i, y_i)$, we determine preliminary action boundaries by identifying the nearest background points immediately preceding and succeeding the annotated point, denoted by $\beta_i = [b^s_i, b^e_i]$. If point $t_i$ belongs to action class $c$ (i.e., $y_i[c] = 1$), the objective is to estimate the boundaries of the $i$-th action instance using signal $\tilde{P}_L[t, c]$ within the interval $\beta_i$. Within interval $\beta_i$ and within distance $\delta d_i$ from the annotated point $t_i$, we locate the snippet $t^{\star}_i$ with the peak probability of class $c$. Here, $d_i$ is the duration of $\beta_i$ and $\delta$ is a hyper-parameter.

\vspace{-5pt}
\begin{equation}
   t^{\star}_i = \operatorname*{argmax}_{t}(\tilde{P}_L[t,c]) \text{ \ for \ } t \in (\beta_i \cap [t_i-\delta d_i,t_i+\delta d_i]). 
\end{equation}

The intuition behind selecting the peak point $t^{\star}_i$ is that this point is the most representative snippet of class $c$ in the vicinity of point $t_i$. The point $t^{\star}_i$ is treated as the mean of the uniform and the Gaussian distributions. For the $i$-th action instance, the signal $\tilde{P}_L[t, c]$ is set to zero outside the interval $\beta_i$. We fit a Gaussian distribution centered at $t^{\star}_i$ to $\tilde{P}_L[t, c]$ for each action instance. Gaussian distribution is defined as follows where $t$, $\mu$, and $\sigma$ represent the temporal axis, mean, and standard deviation.

\begin{equation}
G(t, \mu, \sigma) = \frac{1}{\sigma\sqrt{2\pi}} e^{-\frac{1}{2}\left(\frac{t - \mu}{\sigma}\right)^2}
\end{equation}

The Gaussian distribution can be uniquely defined for the $i$-th action instance as $G(t, t^{\star}_i, \sigma_i)$ by estimating the standard deviation $\sigma_i$. An upper bound $u_b$ and a lower bound $l_b$ are estimated for $\sigma_i$ with respect to boundaries of $\beta_i$.
\vspace{-5pt}
\begin{equation}
u_b = \max(t^{\star}_i - b^s_i, b^e_i - t^{\star}_i), \ l_b = 10^{-6}.
\end{equation}

Thus, the objective is to find the optimal $\sigma_i$ within range $[l_b, u_b]$ to fit Gaussian distribution $G(t, t^{\star}_i, \sigma_i)$ to probability signal $\tilde{P}_L[t, c]$. We address this optimization problem by minimizing the following MSE loss using Brent's method with the bounded variant \cite{brent2013algorithms}.

\vspace{-5pt}
\begin{equation}
L^{\text{G-fit}}_{\text{MSE}}= \sum_{t \in \beta_i} \left( \alpha \cdot G(t, t^{\star}_i, \sigma_i) - \tilde{P}_L[t, c] \right)^2.
\end{equation}

$\alpha$ is a scale factor equal to $\tilde{P}_L[t^{\star}_i,c] / G(t^{\star}_i, t^{\star}_i, \sigma_i)$. Brent's method \cite{brent2013algorithms} is a root-finding algorithm that iteratively adjusts the sigma $\sigma_i$ within specified bounds $l_b$ and $u_b$ to find an optimal standard deviation for the Gaussian component. The same process is applied to find an ideal width $\omega_i$ for the uniform component.

\vspace{-3pt}
\begin{equation}
L^{\text{U-fit}}_{\text{MSE}}= \sum_{t \in \beta_i} \left(U(t^{\star}_i, \omega_i)  - \tilde{P}_L[t,c] \right)^2.
\end{equation}

The linear combination of parameters $\sigma_i$ and $\omega_i$ defines the final interval duration $\Delta_i$ for the $i$-th action where $\Delta_i = \gamma_1 \sigma_i + \gamma_2 \omega_i$. The duration $\Delta_i$ defines the estimated interval $I_i=[t^{\star}_i-\Delta_i, t^{\star}_i+\Delta_i]$. For each video, the pseudo-labels set includes the annotated point $t_i$, the predicted sigma $\sigma_i$, the estimated interval $I_i$, and the label $y_i$, as below:

\vspace{-5pt}
\begin{equation}
\label{PS_ADM}
 \Psi = \{(t_i, \sigma_i, I_i, y_i)\}_{i=1}^{N_{\text{act}}}  \text{ \ where \ } I_i=[t^{\star}_i-\Delta_i, t^{\star}_i+\Delta_i].
\end{equation}

\subsubsection{The Main Model: ADM-Loc}
\label{main-model}

The backbone of the main model is a multi-scale transformer (described in \ref{Backbone}). The model is supervised with the pseudo-labels set $\Psi = \{(t_i, \sigma_i, I_i, y_i)\}_{i=1}^{N_{\text{act}}}$ generated by actionness distribution modeling in eq. \ref{PS_ADM}. The main model is trained with the losses in eq. \ref{total-loss} as well as two additional losses introduced in this section.

\textbf{Learning boundary snippets}. The $\mathcal{L}_{\text{Act}}$ loss (eq. \ref{act-loss}) supervises the learning of probability signal $\hat{P_l}$ for the $i$-th action instance only within interval $I_i$ which is merely an estimation of the the action boundaries. It is probable that the interval $I_i$ fails to encompass snippets near the action boundaries, which are often ambiguous and include transitional movements. Nevertheless, the model needs to classify these boundary snippets as part of the action to generate complete action proposals during testing. Although the action probabilities at these boundary snippets might be lower compared to the more representative action snippets, it remains essential for the model to differentiate these boundary snippets from the background. We impose this by comparing the probability signal $\hat{P_l}$ with a Gaussian kernel, reinforcing the consistency of action classification probabilities for the entire duration of action. The probability signal predicted by the first level of the feature pyramid, denoted as $\hat{P_1} \in \mathbb{R}^{T_1 \times C+1}$, exhibits the highest variability in action probability predictions due to its small receptive field. As a result, this signal particularly benefits from being compared against a Gaussian kernel to stabilize these fluctuations.

\textbf{Standard deviation prediction.} The extracted feature sequence from the first level of the pyramid is denoted by $Z^1 \in \mathbb{R}^{T_1 \times D}$. For the $i$-th action instance, $K$ features are sampled from $Z^1$ within pseudo-label interval $I_i$ and fed to a regression head to predict the standard deviation $\Tilde{\sigma}_{i}$. The regression head consists of temporal convolutions, layer normalization, and the sigmoid function to predict the value of $\Tilde{\sigma}_{i}$ between $[0,1]$. We use the values of $\{\sigma_i\}_{i=1}^{N_\text{act}}$ from the pseudo-labels to determine parameter $K$ and re-scale the predicted $\Tilde{\sigma}_{i}$. This prediction is supervised using an MSE loss, which measures the discrepancy between the predicted and pseudo-label standard deviations. 

\vspace{-3pt}
\begin{equation}\mathcal{L}^{\sigma}_{\text{MSE}} = \frac{1}{N_{\text{act}}} \sum_{i=1}^{N_{\text{act}}} (\sigma_{i}-\Tilde{\sigma}_{i})^2.
\label{MSE-sigma-loss}
\end{equation}

\begin{table*}[th!]
\centering
\resizebox{2\columnwidth}{!}{%
\begin{tabular}{c l  c c c c c  c | c c c c}
 \hline
 \multicolumn{1}{c}{\multirow{3}{*}{Group}} &
\multicolumn{1}{l}{\multirow{3}{*}{Method}} &
 \multicolumn{6}{c}{THUMOS'14} &
  \multicolumn{4}{|c}{ActivityNet-v1.2} \\ \cline{3-8} \cline{9-12}

\multicolumn{1}{c}{}  &  \multicolumn{1}{c}{} &  \multicolumn{5}{c}{mAP@IoU (\%)}& mAP-AVG& 
\multicolumn{3}{c}{mAP@IoU (\%)} & mAP-AVG\\ \cline{3-8} \cline{9-12}

&   &  0.3	&0.4	&0.5	&0.6	&0.7 & (0.1:0.7) & 0.5 & 0.75 & 0.95 & (0.5:0.95) \\
\hline

 \multicolumn{1}{c}{\multirow{15}{*}{WS}} & ASL\cite{ma2021weakly} &51.8  & - & 31.1 & - & 11.4 & 40.3 & 40.2 & - & - & 25.8 \\

&CoLA \cite{zhang2021cola} & 51.5 & 41.9 & 32.2 & 22.0 & 13.1  & 40.9 &   42.7 &  25.7 &  5.8 & 26.1 \\


& AUMN \cite{luo2021action} & 54.9 &  44.4 & 33.3 &20.5  &9.0   & 41.5 & 42.0 &  25.0 & 5.6 &  25.5 \\


& FTCL \cite{gao2022fine} & 55.2 & 45.2 & 35.6 & 23.7 & 12.2  & 43.6  & - & - & - & -\\

& UGCT \cite{yang2021uncertainty}  &  55.5 & 46.5 & 35.9 & 23.8 & 11.4  & 43.6 &  41.8 &  25.3 &  5.9 &  25.8  \\ 

&CO2-Net \cite{hong2021cross}& 58.2 & 47.1 & 35.9 & 23.0 & 12.8 & - & 43.3 &  26.3 &  5.2 & 26.4 \\

 &  D2-Net \cite{narayan2021d2} &   52.3  &  43.4  & 36.0 & - & -   & - & 42.3 &  25.5 &  5.8 &  26.0 \\

& ASM-Loc\cite{he2022asm}   &   57.1 & 46.8 & 36.6 & 25.2 & 13.4  &   45.1 & - & - & - & - \\ 

& RSKP\cite{huang2022weakly}  & 55.8 & 47.5 & 38.2 & 25.4 & 12.5 &   45.1& - & - & - & -\\

 & TS\cite{wang2023two}  & 60.0 & 47.9 & 37.1 & 24.4 & 12.7 & 46.2& - & - & - & -\\

& DELU\cite{chen2022dual} & 56.5 & 47.7 & 40.5 & 27.2 & 15.3  &  46.4 & 44.2 & 26.7 & 5.4 & 26.9 \\

& P-MIL \cite{ren2023proposal} & 58.9 & 49.0 & 40.0 & 27.1 & 15.1  &  47.0 &  44.2 & 26.1 & 5.3 & 26.5 \\

& Zhou \textit{et al.} \cite{zhou2023improving} & 60.7 & 51.8 & 42.7 & 26.2 & 13.1 & 48.3 & - & - & - & - \\

& PivoTAL \cite{rizve2023pivotal} & 61.7 & 52.1 & 42.8 & 30.6 & 16.7  & 49.6 & - & - & - & -\\

\hline
\hline

\multicolumn{1}{c}{\multirow{6}{*}{PS}} & SF-Net \cite{ma2020sf} &   52.8 & 42.2 & 30.5 & 20.6 & 12.0  & 41.2 & 37.8 & - & - & 22.8 \\ 

& Je \textit{et al.} \cite{ju2021divide}  & 58.1 & 46.4 & 34.5 & 21.8 & 11.9 & 44.3& - & -& -& -\\

& PTAL \cite{ju2020point}  &  58.2 & 47.1 & 35.9 & 23.0 & 12.8 & - & -& -& -& -\\ 

& BackTAL \cite{yang2021background}  &  54.4 & 45.5 & 36.3 & 26.2 & 14.8 &  - & 41.5 & 27.3  & 4.7 & 27.0\\ 

& Lee \textit{et al.} \cite{lee2021learning}  & 64.6 & 56.5 & 45.3 & 34.5 & 21.8  & 52.8 & 44.0 & 26.0 & 5.9  & 26.8 \\ 

& \textbf{Our ADM-Loc} &  \textbf{71.5} & \textbf{64.7} & \textbf{56.0} & \textbf{43.2} & \textbf{31.3} &  \textbf{60.2} & \textbf{44.5} & \textbf{28.0} &  \textbf{6.2} & \textbf{27.9} \\

\hline 
\end{tabular}}
\caption{Comparison with weakly-supervised (WS) and point-supervised (PS) methods on THUMOS'14 and ActivityNet-v1.2. The results are reported in terms of mAP (\%) at different tIoU thresholds. The bold numbers show the best results. }
\label{SOTA-thumos}
\end{table*}

\textbf{Gaussian imposition}. The set $S_c$ denotes the set of action classes that occur in a given video. A Gaussian kernel is defined to represent the $i$-th action instance formulated as follows where $t_i$ is the annotated point and $\Tilde{\sigma}_{i}$ is the predicted standard deviation. 

\begin{equation}
 G_i(t, t_i, \Tilde{\sigma}_{i}) = e^{-\frac{1}{2}\left(\frac{t - t_i}{\Tilde{\sigma}_{i}}\right)^2}.   
\end{equation}

For each action class $c \in S_c$, we mix the Gaussian kernels of all action instances belonging to class $c$, as follows. 

\begin{equation}
G^c(t) = \max \left\{ G_i(t, t_i, \Tilde{\sigma}_{i}) | \ i \in [1, N_{\text{act}}],  y_i[c]=1 \right\}.
\end{equation}

The alignment between the probability signal $\hat{P}_1[t,c]$ and the Gaussian kernel $G^c(t)$ is supervised using the following MSE loss. 

\begin{equation}
\mathcal{L}^G_{\text{MSE}}= \frac{1}{T_1 |S_c| } \sum_{c \in S_c} \sum_{t=1}^{T_1} \left( G^c(t) - \hat{P}_1[t,c] \right)^2.
\label{MSE-G-loss}
\end{equation}

\textbf{Pseudo-label sampling}. We incorporate a pseudo-label sampling strategy during the training process for $\mathcal{L}_{\text{Act}}$ loss by selecting the snippets around the annotated points within a radius hyper-parameter $r_s$ and inside the boundaries of pseudo-labels. The motivation for this sampling is to reduce the likelihood of training the model on false positives. During the pseudo-label generation, the background frames that are erroneously classified as actions constitute the false positives. These are more likely to occur at the boundaries of the pseudo-labels.

\textbf{Joint training}. The total loss for the main model is a weighted combination of the following losses where $\lambda_{\star}$ are determined through empirical analysis.

\begin{align}
  \footnotesize
  \mathcal{L}_{\text{Total}} =  &\lambda_{\text{MIL}}\mathcal{L}_{\text{MIL}} + \lambda_{\text{Act}} \mathcal{L}_{\text{Act}} + \lambda_{\text{BG}} \mathcal{L}_{\text{BG}}  \nonumber \\
  &+ \lambda_{\text{G}}\mathcal{L}^G_{\text{MSE}} + \lambda_{\sigma} \mathcal{L}^{\sigma}_{\text{MSE}} .  
    \normalsize
\end{align}

\textbf{Inference}. The action categories are identified using the video-level scores. The action proposals are predicted from all pyramid levels by applying thresholds to the snippet-level action scores $\hat{P_l}$ for each level $l$ for the predicted classes and merging consecutive candidate segments. Each proposal is assigned a confidence score based on its outer-inner-contrast score \cite{lee2020background}. Finally, the non-maximum suppression (NMS) is used to eliminate overlapping proposals.

\section{Experiments}

\subsection{Experimental Setting}

\textbf{Datasets}. THUMOS14 \cite{jiang2014thumos} comprises untrimmed videos across 20 unique categories. In line with prior work \cite{lee2021learning,he2022asm}, we use the 200 videos in the validation set for training and the 213 videos in the testing set for evaluation. The average number of action instances per video is $15.5$. ActivityNet-v1.2 is a large-scale dataset containing $9,682$ videos that includes 100 complex everyday activities. The average number of action instances per video is $1.5$. Consistent with previous work, our model is trained using the training set and evaluated using the validation set \cite{lee2021learning,he2022asm}. 

\textbf{Evaluation metric}. The Mean Average Precision (mAP) under different Intersection over Union (IoU) thresholds is utilized as the evaluation metric, wherein the Average Precision (AP) is computed for each action class. On ActivityNet-v1.2 \cite{caba2015activitynet}, IoU thresholds range from $0.5$ to $0.95$ in increments of $0.05$. As for THUMOS14 \cite{jiang2014thumos}, they range from $0.1$ to $0.7$ in increments of $0.1$.

\textbf{Implementation details}. For feature extraction, we use two-stream I3D \cite{carreira2017quo} on both datasets. We fed $16$ consecutive frames as the input to the visual encoder, using a sliding window with stride $4$ on THUMOS14 and stride $16$ on ActivityNet-v1.2.  Our multi-scale transformer model is trained with Adam \cite{kingma2014adam} and linear warm-up \cite{liu2020understanding} with the learning rate of $10^{-4}$. Model EMA \cite{huang2017snapshot} is implemented to further stabilize the training. The number of epochs and warm-up epochs are set to $100$ and $10$ on THUMOS14, and $50$ and $5$ on ActivityNet-v1.2. The batch sizes are set to $3$ on THUMOS14, and $64$ on ActivityNet-v1.2. The input length is set to $2,304$ for THUMOS14 and to $192$ for ActivityNet-v1.2, using padding, random sampling and linear interpolation. To employ local self-attention, the window lengths are set to $19$ and $7$ on THUMOS14 and ActivityNet-v1.2, respectively. The number of pyramid levels is set to $L=4$ and the down sampling ratio $\theta$ is set to $2$. The annotation augmentation radius $r_a$ and the pseudo-label sampling radius $r_s$ are set to $2$. The parameters $r_a$ and $r_s$ are defined on the feature grid, representing the distance in terms of the number of features. At inference, the full sequence is fed into the model without sampling. \footnote{The source code will be released upon acceptance of the paper.}

\subsection{Comparison with State-of-the-art Methods}

Table \ref{SOTA-thumos} shows a detailed comparison with the leading methods on THUMOS'14 and ActivityNet-v1.2 datasets.



\textbf{Results on THUMOS’14}: Our model significantly outperforms other point-supervised methods, achieving an average mAP improvement of $7.4\%$. Notably, this enhancement is nearly $10\%$ at the most stringent IoU threshold of $0.7$. Moreover, our model shows a significant gain of $10.6\%$ average mAP increase over weakly-supervised methods, despite using only slightly more annotations. The mAP at the $0.7$ IoU threshold is almost double that of its weakly-supervised counterparts.

\textbf{Results on ActivityNet-v1.2}. Our model outperforms the state-of-the-art weakly and point-supervised methods in terms of mAP, consistently across all the IoU thresholds.

\subsection{Ablation Studies}

\textbf{Quality of pseudo-labels}. In Table \ref{PL-generation-quality}, $\alpha$ represents the ratio of the number of generated proposals to the ground-truth instances in the training set (validation set of THUMOS'14). The first row shows the quality of the generated action proposals on the training set using the base model (section \ref{base-mode}). As shown in the table, the average mAP of these proposals is only $63.8\%$, and the number of predicted proposals is 12 times the number of ground-truth instances ($\alpha=12$). This indicates that a large number of proposals are redundant and overlapping, making them unsuitable to be used as pseudo-labels. Ideally, there should be a one-to-one correspondence between the pseudo-labels and action instances. The second row demonstrates the quality of pseudo-labels generated using our proposed Actionness Distribution Modeling (ADM), as detailed in section \ref{ADM-section}. Noticeably, the mAP at the highest IoU of $0.7$ is almost doubled compared to the base proposals. Furthermore, ADM generates exactly one proposal (pseudo-label) for each annotated point ($\alpha=1$).

\begin{table}[h!]
\centering
\resizebox{\columnwidth}{!}{%
\begin{tabular}{c | c | c | c  }
 \hline
\multicolumn{1}{c|}{\multirow{2}{*}{Method}} & \multicolumn{1}{c|}{\multirow{2}{*}{$\alpha$}} & 
\multicolumn{1}{c|}{\multirow{2}{*}{mAP@0.7 (\%)}} & 
mAP-AVG (\%)\\ 
& &  & (0.1:0.7) \\
\hline
Base Model (ours) & $\sim$ 12  & 20.1 &   63.8 \\
\textbf{ADM-Loc} (ours) & $\sim$ 1 & \textbf{38.4} &  \textbf{76.4}\\
\hline 
\end{tabular}}
\caption{Analysis of the generated pseudo-labels on the \textbf{validation set} of THUMOS'14. $\alpha$ represents the ratio of the number of generated proposals to the ground-truth instances.
}
\label{PL-generation-quality}
\end{table}

\textbf{Impact of pseudo-labels}. Table \ref{PL-impact} shows the impact of supervision in the base model. The first row shows supervision with only points without augmentation ($r_a=0$), achieving the lowest results. We also compare the performance of the base model when supervised with the augmented points ($r_a=2$) versus the sampled pseudo-labels ($r_s=2$). Note that the radius (for both $r_s$ and $r_a$) on level $l$ of the pyramid is $r_{\ast} \cdot \theta^l = 2^{l+1}$, for $\theta=2$ and $r_{\ast}=2$. Therefore, the radius can be as large as $32$ for level $l=4$. The model trained with augmented points selects all snippets within the radius as positive samples, even if the action duration is much shorter than the radius. In contrast, the pseudo-labels effectively limit the positive samples to estimated action boundaries. This results in a performance gain of $5.7\%$ average mAP.

 

\begin{table}[h!]
\centering
\resizebox{\columnwidth}{!}{%
\begin{tabular}{ c | c | c c c | c }
\hline
 \multirow{2}{*}{Supervision} & \multirow{2}{*}{Radius}  & \multicolumn{3}{c|}{mAP@IoU (\%)}& mAP-AVG (\%) \\ \cline{3-6}
& & 0.3	& 0.5 & 0.7 & (0.1:0.7) \\ 
\hline
Only Points & $r_a=0$ & 29.4 & 16.1 & 5.4 & 23.7 \\
\hline
Augmented Points & $r_a=2$ &  65.6 & 45.9 & 20.1 & 53.2 \\
\hline
Pseudo-labels  & $r_s=2$ & \textbf{69.7} & \textbf{55.0}  &  \textbf{29.0} & \textbf{58.9}\\
\hline
\end{tabular}}
\caption{Impact of the generated pseudo-labels in the base model on THUMOS'14. $r_a$ is the annotation augmentation radius, and $r_s$ is the pseudo-label sampling radius.}
\label{PL-impact}
\end{table}


\textbf{Impact of the backbone network}. Table \ref{backbone-impact} demonstrates the impact of the backbone multi-scale transformer architecture in ADM-Loc (the main model). Parameter $l$ denotes the number of feature pyramid levels. The pseudo-label sampling radius $r_s$ is set to $2$. As shown in the table, the highest average mAP is achieved when $l=4$, which is comparable to the results for $l=5$.


\begin{table}[h!]
\centering
\resizebox{0.9\hsize}{!}{
\begin{tabular}{ c | c c c c c c }
\hline
Levels & $l=1$ & $l=2$ & $l=3$ & $l=4$ & $l=5$ \\
\hline
mAP@0.7 & 24.3 & 29.9 & 28.5 & \textbf{31.3} & 30.5 \\
\hline
mAP-AVG & 56.0 & 59.6 & 59.5 & \textbf{60.2} & \textbf{60.4} \\
\hline
\end{tabular}}
\caption{Impact of the number of pyramid levels (denoted by $l$) in ADM-Loc backbone on THUMOS'14. }
\label{backbone-impact}
\end{table}

\textbf{Impact of pseudo-label sampling}. Table \ref{radius-impact} demonstrates the impact of the pseudo-label sampling strategy with different sampling radius $r_s$ on ADM-Loc (the main model). $r_s= \infty$ indicates no sampling. As indicated in the table, using a sampling radius of $r_s=2$ results in a $3.2\%$ improvement in average mAP compared to the scenario with no sampling ($r_s= \infty$). This is because pseudo-label sampling decreases the chance of training the model on false positives with the $\mathcal{L}_{\text{Act}}$ loss (see eq. \ref{act-loss}). During the pseudo-label generation, the background frames that are erroneously classified as actions constitute the false positives. These are more likely to occur at the boundaries of the pseudo-labels.

\begin{table}[h!]
\centering
\resizebox{0.8\columnwidth}{!}{%
\begin{tabular}{ c | c c c | c }
\hline
\multirow{2}{*}{Radius ($r_s$)} & \multicolumn{3}{c|}{mAP@IoU (\%)}& mAP-AVG \\ \cline{2-5}
 & 0.3	& 0.5 & 0.7 & (0.1:0.7) \\
\hline
1 & 70.7 & 55.1 & 28.4 & 59.1\\
2 & \textbf{71.5} & \textbf{56.0} & \textbf{31.3} & \textbf{60.2}\\
4 & 70.8 & 54.8 & 30.1 & 59.8 \\
$\infty$ & 68.1 & 51.9 & 28.2  & 57.0\\
\hline
\end{tabular}}
\caption{Impact of the pseudo-label sampling radius $r_s$ in ADM-Loc on THUMOS'14 where $r_s= \infty$ means no sampling. }
\label{radius-impact}
\end{table}


\textbf{Impact of the proposed losses}. Table \ref{impact-losses} demonstrates the impact of the proposed losses $\mathcal{L}^{\sigma}_{\text{MSE}}$ (eq. \ref{MSE-sigma-loss}) and $\mathcal{L}^{G}_{\text{MSE}}$ (eq. \ref{MSE-G-loss}) in ADM-Loc. All experiments in this table are also supervised with $\mathcal{L}_{\text{MIL}}, \mathcal{L}_{\text{Act}}, \mathcal{L}_{\text{BG}}$ (eq. \ref{eq.MIL}, \ref{bg-loss}, \ref{act-loss}) losses. The pseudo-label sampling radius $r_s$ is set to 2. As indicated in the table, the implementation of both losses has led to a performance gain of $1.3\%$ in average mAP and $2.3\%$ in mAP at tIoU$=0.7$.

\begin{table}[th!]
\centering
\resizebox{\columnwidth}{!}{%
\begin{tabular}{ c  c | c c c | c }
\hline
 \multicolumn{2}{c|}{Proposed Losses}  & \multicolumn{3}{c|}{mAP@IoU (\%)}& mAP-AVG \\ 
\hline
$\mathcal{L}^{\sigma}_{\text{MSE}}$ & $\mathcal{L}^{G}_{\text{MSE}}$ &  0.3	& 0.5 & 0.7 & (0.1:0.7) \\
\hline
 \ding{55} & \ding{55} &  69.7 & 55.0  &  29.0 & 58.9\\
 \ding{55} &  $\checkmark$ & 70.6 & 55.7 & 30.0 & 59.8 \\
 $\checkmark$ &  $\checkmark$ & \textbf{71.5} & \textbf{56.0} & \textbf{31.3} & \textbf{60.2}\\
\hline
\end{tabular}}
\caption{Impact of the proposed losses in ADM-Loc on THUMOS'14.}
\label{impact-losses}
\end{table}

\subsection{Qualitative Results}

\begin{figure}[t!]
    \centering
    \includegraphics[width=\linewidth]{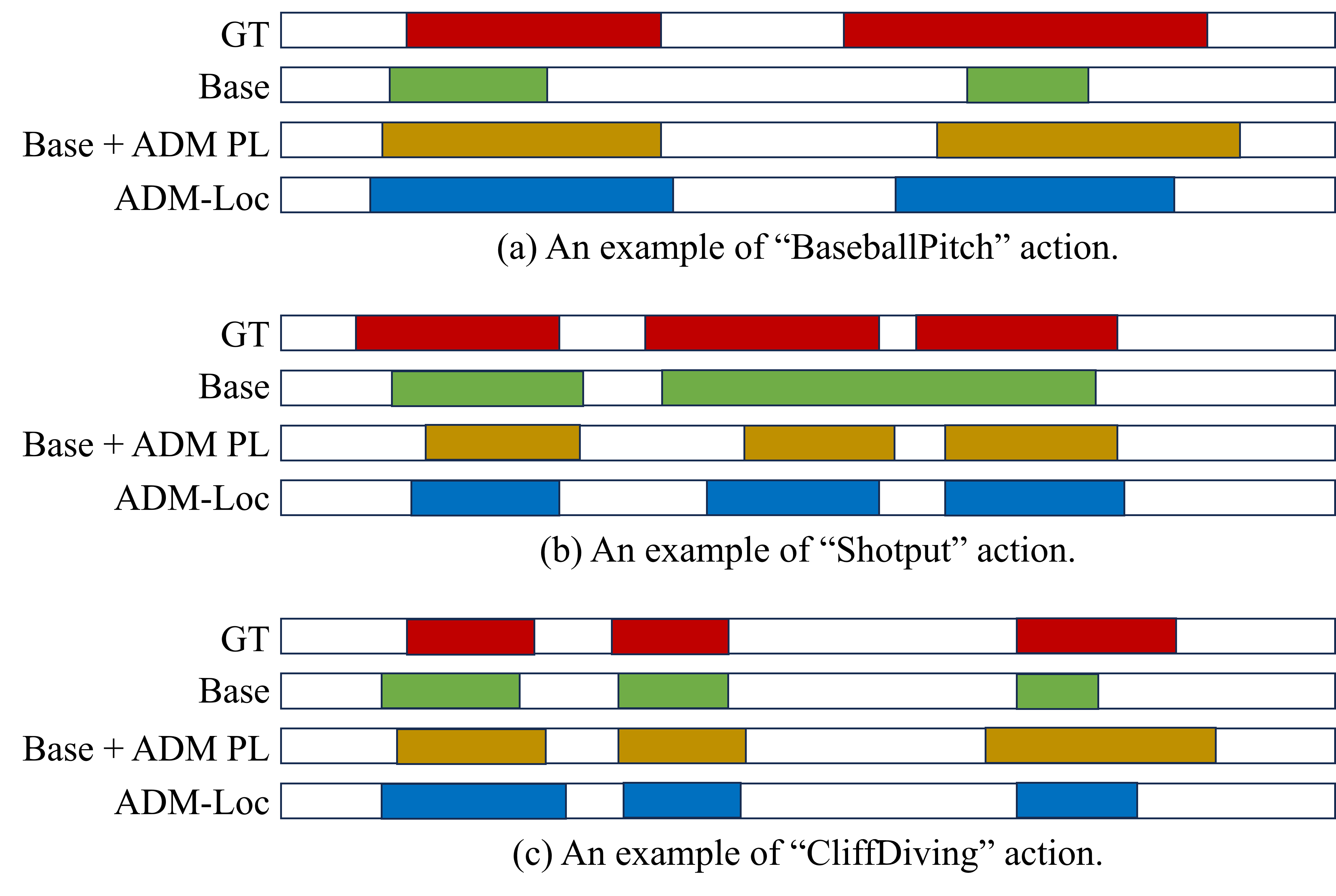}
    \caption{Qualitative results on THUMOS'14 for three different action classes. The ground-truth instances are highlighted in red. The detection results are displayed from: (1) the base model supervised with point-level annotations (green), (2) the base model supervised with our generated pseudo-labels (brown), (3) our ADM-Loc framework (blue).}
    \label{fig:visual-res}
\end{figure}

Fig.~\ref{fig:visual-res} presents the qualitative results of our model in different stages: (1) the base model supervised with point-level annotations (section \ref{base-mode}), (2) the base model supervised with pseudo-labels generated by ADM (section \ref{PLG-ADM-section}), and (3) our full ADM-Loc framework (section \ref{main-model}). This figure demonstrates that ADM-Loc partially addresses misalignments between actual instances and proposals in the base model, such as the incomplete localization of the action `Baseball Pitch' (part a), and the over-complete localization of the action `Shotput' (part b). Furthermore, in some cases, the base model supervised with pseudo-labels generates over-complete proposals (such as the last action instance of `CliffDiving' in part c), which are adjusted in ADM-Loc by modeling action boundary snippets.

\section{Conclusion}

We propose ADM-Loc, a novel point-supervised framework that employs a self-training scheme to generate high-quality pseudo-labels, providing additional supervision during training. Our approach for pseudo-label generation models the action classification probabilities for each action instance in the video. It avoids reliance on arbitrary thresholding, estimates a single action proposal per action instance, and demonstrates robustness to inconsistencies in action classification probabilities. Furthermore, we propose modeling action boundary snippets by enforcing consistency in action classification scores during training, guided by our designed loss functions. ADM-Loc surpasses state-of-the-art point-supervised methods on both THUMOS’14 and ActivityNet-v1.2 datasets.

\section{Acknowledgment}
This material is based upon work supported by the National Science Foundation under award number 2041307.

\section{Appendix}

\subsection{Temporal Action Detection Error Analysis}

\begin{figure*}[t!]
    \centering
    \begin{subfigure}[b]{\textwidth}
        \centering
        \includegraphics[width=0.7\textwidth]{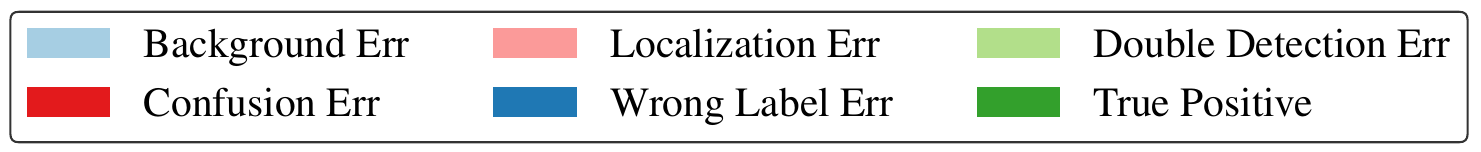}
    \end{subfigure}
    \vspace{1em}
    \begin{subfigure}[b]{0.32\textwidth}
        \centering
        \includegraphics[width=\textwidth]{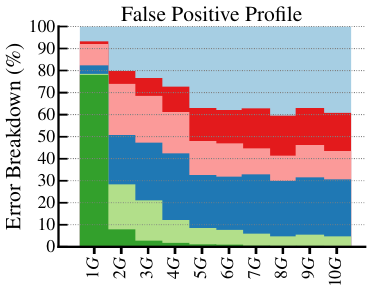}
        \caption{ActionFormer \cite{zhang2022actionformer} (Fully-supervised)}
    \end{subfigure}
    \hfill 
    \begin{subfigure}[b]{0.32\textwidth}
        \centering
        \includegraphics[width=\textwidth]{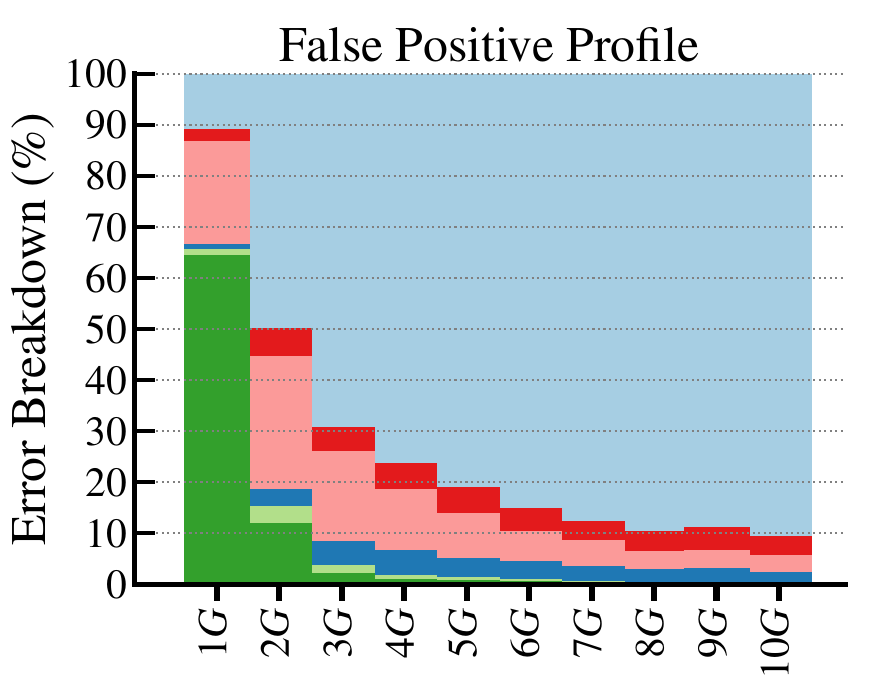}
        \caption{Our ADM-Loc (Point-supervised)}
    \end{subfigure}
    \hfill 
    \begin{subfigure}[b]{0.32\textwidth}
        \centering
        \includegraphics[width=\textwidth]{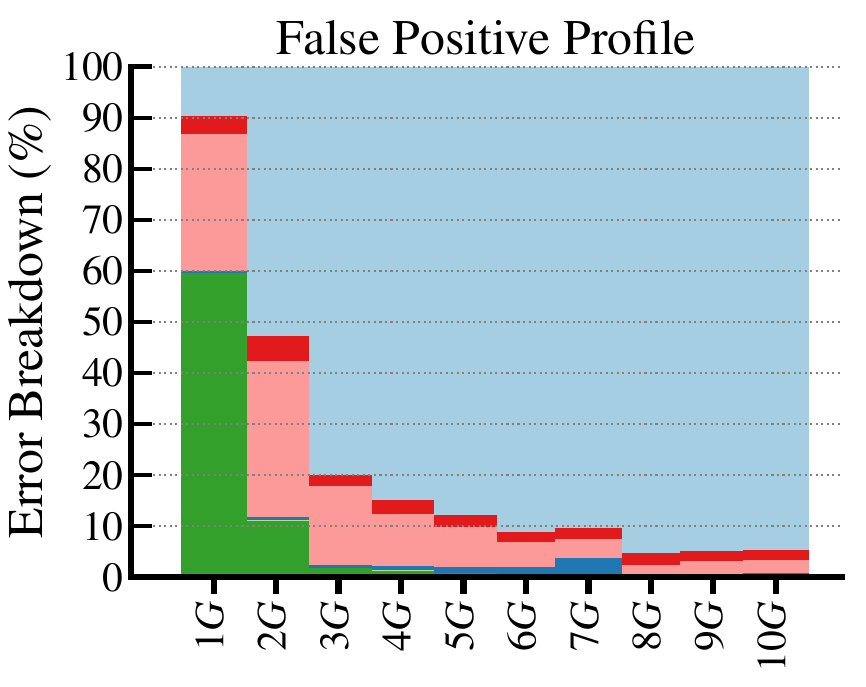}
        \caption{Our base model (Point-supervised)}
    \end{subfigure}
    \caption{False positive (FP) profiling of ActionFormer \cite{zhang2022actionformer} (fully-supervised), our ADM-Loc (point-supervised) and our base model (point-supervised) on THUMOS14 using DETAD \cite{alwassel2018diagnosing}.}
    \label{fig:FP_profiles}
\end{figure*}

To assess the effectiveness and limitations of our ADM-Loc framework, we employ DETAD \cite{alwassel2018diagnosing} for analyzing false negatives (Figure \ref{fig:combined_FN}) and false positives (Figure \ref{fig:FP_profiles}).

\subsubsection{False Negative Analysis}

\begin{figure}[t!]
    \centering
    \begin{subfigure}[b]{\linewidth}
        \includegraphics[width=\linewidth]{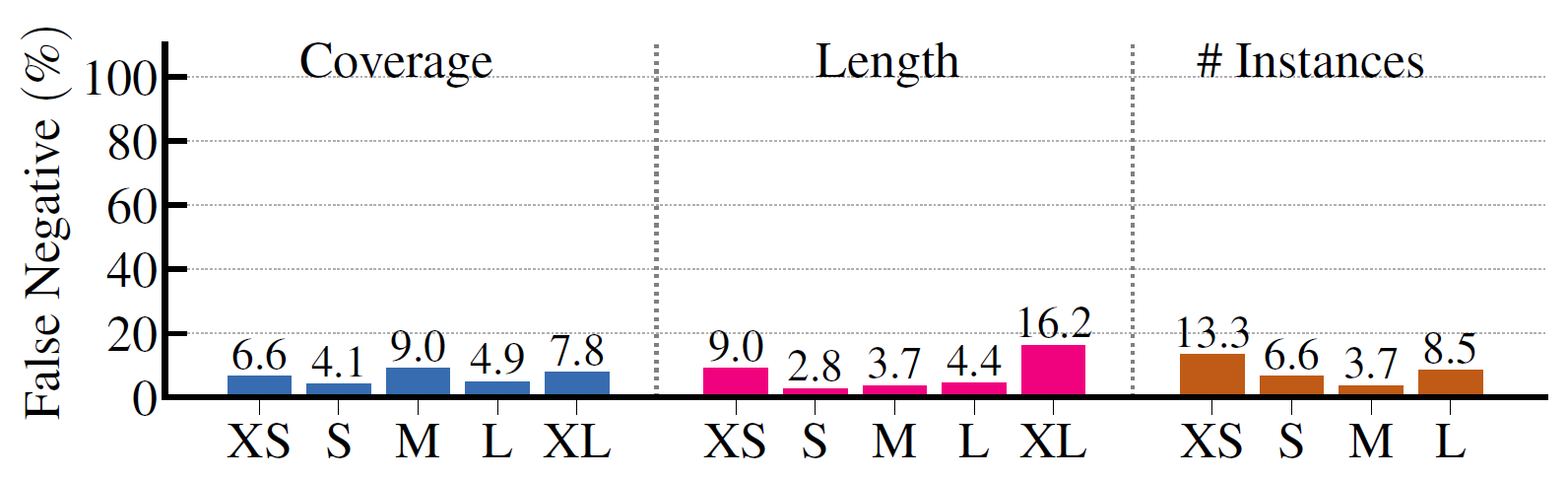}
        \caption{ActionFormer \cite{zhang2022actionformer} (Fully-supervised).}
        \label{fig:ActionFormer_FN}
    \end{subfigure}
    \hfill
    \begin{subfigure}[b]{\linewidth}
        \includegraphics[width=\linewidth]{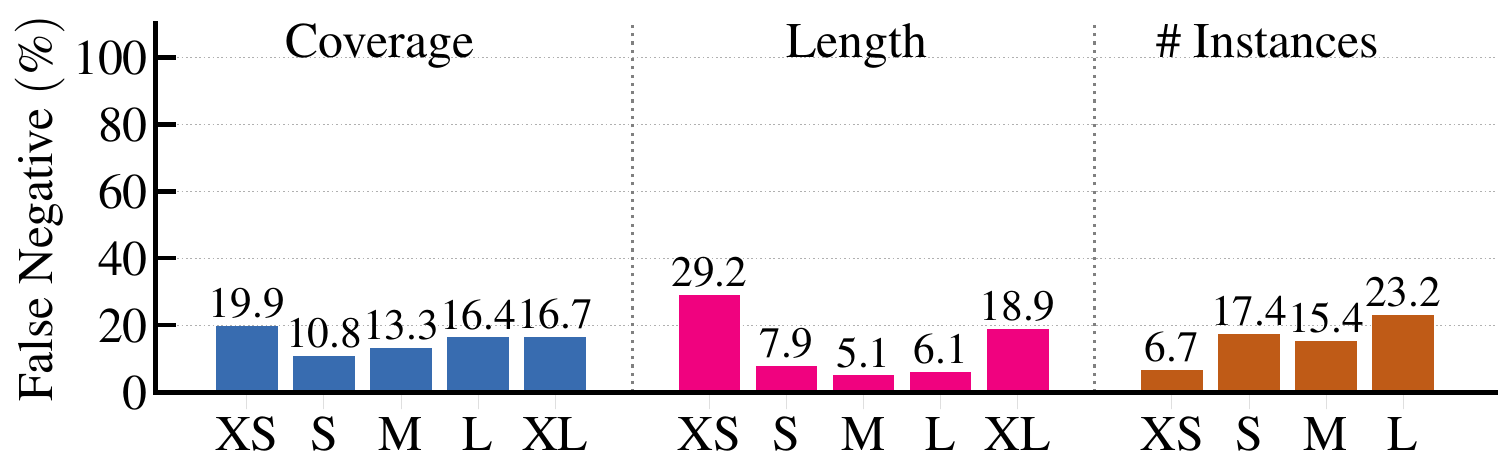}
        \caption{Our ADM-Loc model (Point-supervised).}
        \label{fig:ADMLoc_FN}
    \end{subfigure}
    \hfill
    \begin{subfigure}[b]{\linewidth}
        \includegraphics[width=\linewidth]{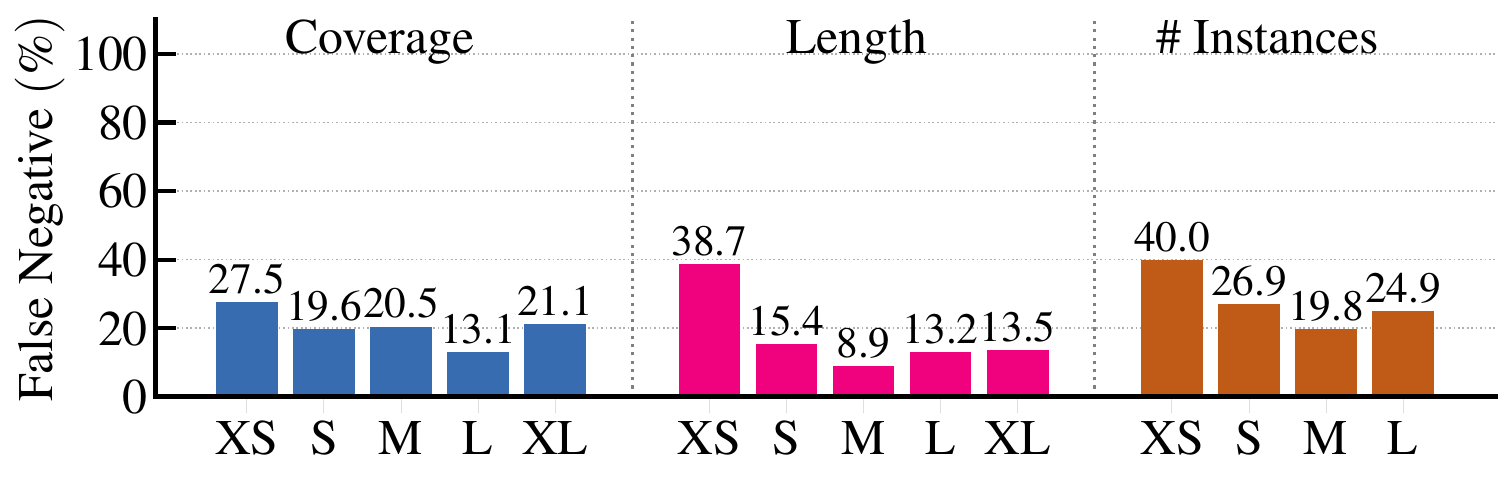}
        \caption{Our base model (Point-supervised).}
        \label{fig:base_FN}
    \end{subfigure}
    \caption{False negative (FN) profiling of ActionFormer \cite{zhang2022actionformer} (fully-supervised), our ADM-Loc (point-supervised) and our base model (point-supervised) on THUMOS14 using DETAD \cite{alwassel2018diagnosing}.}
    \label{fig:combined_FN}
\end{figure}

Figure \ref{fig:combined_FN} illustrates the false negative (FN) profiling across various coverages, lengths, and number of instances. Part (b) of Figure \ref{fig:combined_FN} displays the FN profiling specific to ADM-Loc. The figure reveals that higher false negative rates are associated with action instances characterized by: (1) extremely short or long durations relative to the video length (Coverage (XS) or Coverage (XL)), (2) actions of very short or very long lengths (Length (XS) or Length (XL)), and (3) videos containing a large number of action instances ($\#$Instances (L)). Furthermore, Figure \ref{fig:combined_FN} demonstrates that ADM-Loc (part b) reduces the false negative (FN) rate compared to the base model (part c), except in two cases: Coverage (L) and Length (XL). This is because the base model samples all snippets within the sampling radius for point augmentations, whereas ADM-Loc only samples snippets that fall within the pseudo-label boundaries. To examine the limitations of ADM-Loc relative to fully-supervised methods, FN profiling of ActionFormer\cite{zhang2022actionformer} is provided in Figure \ref{fig:combined_FN} (part a). The most significant FN differences between ActionFormer (part a) and ADM-Loc (part b) are the following cases: Length (XS), $\#$Instances (L). This demonstrates that the annotation of action boundaries is crucial for detecting very short action instances and for accurate detection in videos containing numerous instances.

\subsubsection{False Positive Analysis}
Figure \ref{fig:FP_profiles} presents a detailed categorization of false positive errors and summarizes their distribution. G represents the number of ground truth segments in the THUMOS-14 dataset. This figure indicates that, in comparison with Actionformer (part a), the majority of false positive errors in ADM-Loc (part b) stem from background errors. This occurs because ADM-Loc, as a point-supervised method, lacks access to precise action boundaries. Consequently, background snippets close to action boundaries, sharing characteristics with action instances, may be erroneously detected as actions, resulting in false positives. We also analyze the false positive profiling of ADM-Loc (part b) against the base model (part c) focusing on the top 1G scoring predictions. This comparison reveals that ADM-Loc identifies more true positive instances and exhibits fewer localization and confusion errors. This confirms the effectiveness of ADM-Loc in predicting more precise action boundaries.

\subsection{Distribution of Annotated Points}

In the point-supervision setting, only a single frame per action instance is annotated in the training set. SF-Net \cite{ma2020sf} proposed to simulate point annotations by sampling a single frame for each action instance. The Uniform distribution method randomly selects a frame within the action boundaries of each action, while the Gaussian distribution method does so with respect to a given mean and standard deviation. Typically, the Gaussian distribution is more likely to sample frames closer to the central timestamps of actions, thereby increasing the chances of choosing a more discriminative snippet. In contrast, the Uniform distribution can sample frames from any part of the action, without this central bias. Table \ref{point-distribution} demonstrates that ADM-Loc attains state-of-the-art results with both Uniform and Gaussian point-level distributions on THUMOS'14, indicating its robustness. However, it is observed that the ADM-Loc's performance is lower with the Uniform distribution as compared to the Gaussian distribution. We conjecture this may be attributed to the Uniform distribution's tendency to select less discriminative snippets for point annotation, which can occur anywhere within the action's extent, such as at the boundaries

\begin{table}[h!]
\centering
\resizebox{\columnwidth}{!}{%
\begin{tabular}{c | c | c  c c | c }
 \hline
 \multicolumn{1}{c|}{\multirow{2}{*}{Distribution}}  &  \multicolumn{1}{c|}{\multirow{2}{*}{Method}} & \multicolumn{3}{c|}{mAP@IoU (\%)} & mAP-AVG (\%)\\ \cline{3-6}
  & &  0.3	&0.5	&0.7 & (0.1:0.7) \\
\hline 
\multicolumn{1}{c|}{\multirow{4}{*}{Gaussian}} & \textbf{ADM-Loc} & \textbf{71.5} & \textbf{56.0} & \textbf{31.3} &  \textbf{60.2}\\ 
& Base Model  & 65.6 & 45.9 & 20.1 & 53.2 \\    
 & LACP\cite{lee2021learning}  & 64.6  & 45.3 & 21.8 & 52.8 \\    
& Ju et al. \cite{ju2020point} & 58.2  & 35.9  & 12.8  &  44.8 \\
& SF-Net \cite{ma2020sf} & 47.4 & 26.2 & 9.1 & 36.7 \\
\hline
\multicolumn{1}{c|}{\multirow{4}{*}{Uniform}} & \textbf{ADM-Loc} & \textbf{66.9} & \textbf{45.6} & \textbf{22.3} & \textbf{53.2} \\
& Base Model  & 63.2 & 39.9 & 14.2 & 49.5 \\    
 & LACP \cite{lee2021learning}  & 60.4 & 42.6 &  20.2 & 49.3 \\ 
 & Ju et al. \cite{ju2020point} &  55.6 &  32.3 & 12.3 &   42.9 \\
 & SF-Net \cite{ma2020sf} & 52.0 & 30.2 & 11.8 & 40.5 \\
\hline 
\end{tabular}}
\caption{Performance comparison of ADM-Loc with Uniform and Gaussian point-level distributions on THUMOS'14. }
\label{point-distribution}
\end{table}

\subsection{More Qualitative Results}

\begin{figure*}[t!]
    \centering
    \begin{subfigure}[b]{\linewidth}
        \centering
        \includegraphics[width=\linewidth]{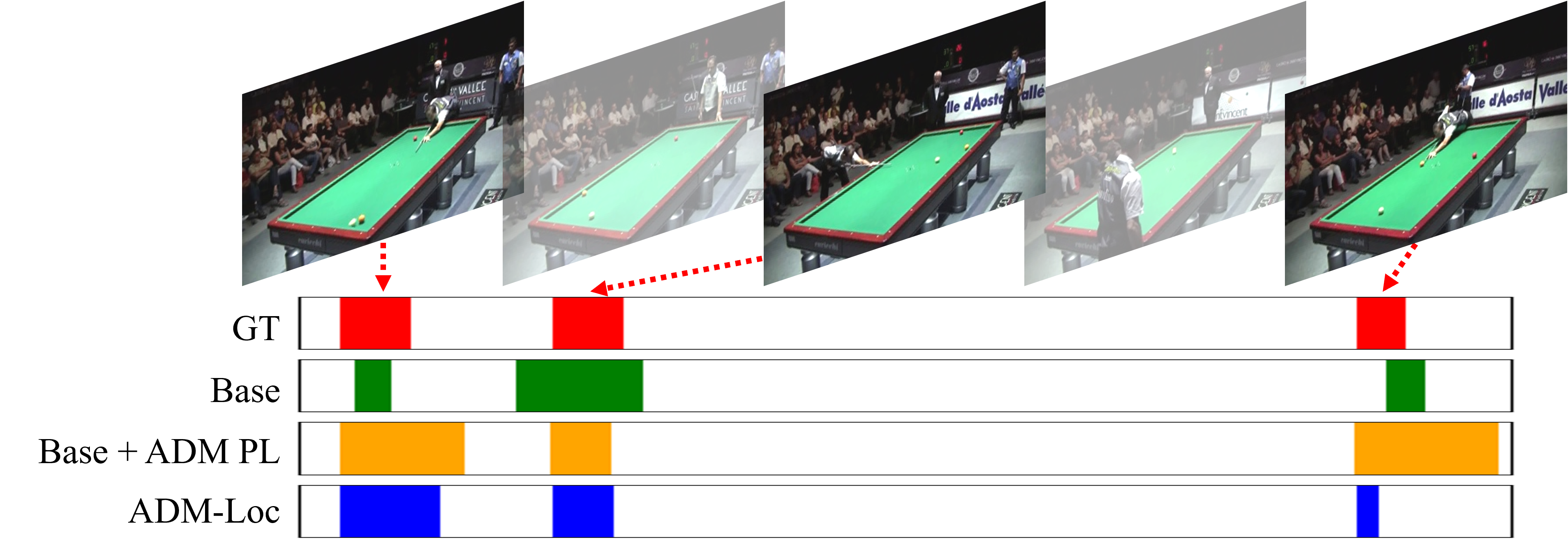}
        \caption{Action ``Billiards".}
    \end{subfigure}
    \hfill 
    \begin{subfigure}[b]{\linewidth}
        \centering
        \includegraphics[width=\linewidth]{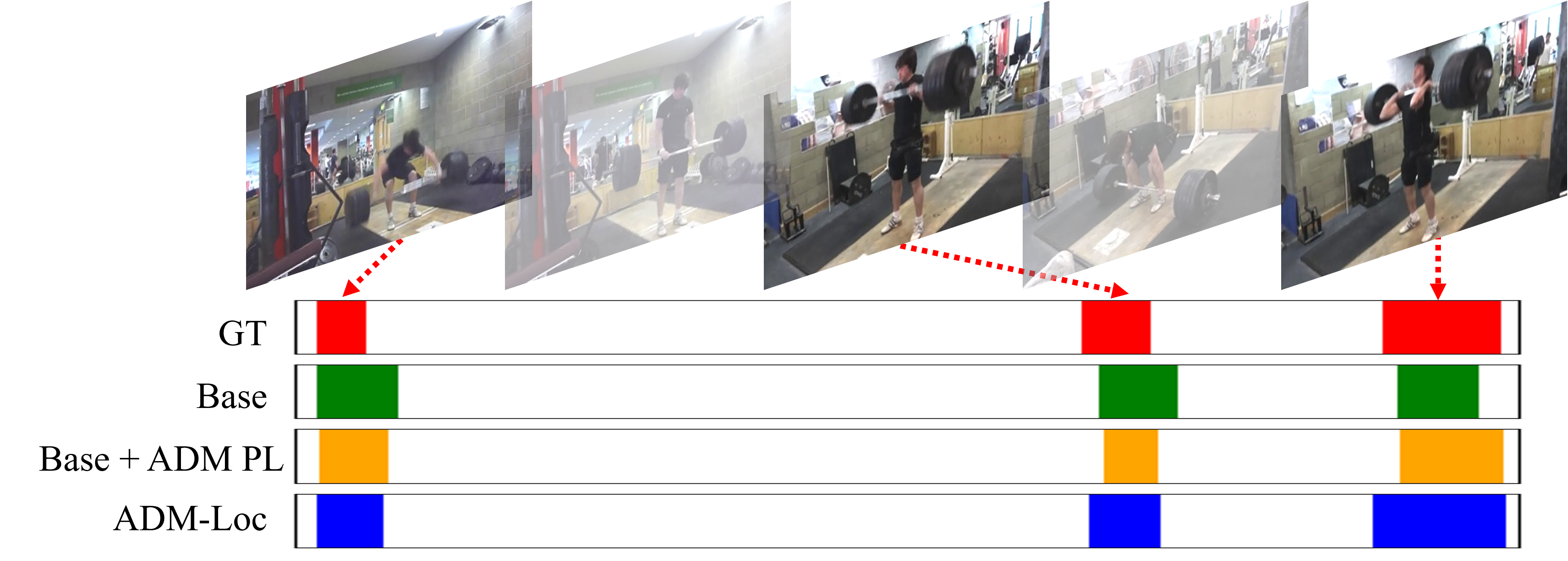}
        \caption{Action ``Clean and jerk".}
    \end{subfigure}
    \hfill 
    \begin{subfigure}[b]{\linewidth}
        \centering
        \includegraphics[width=\linewidth]{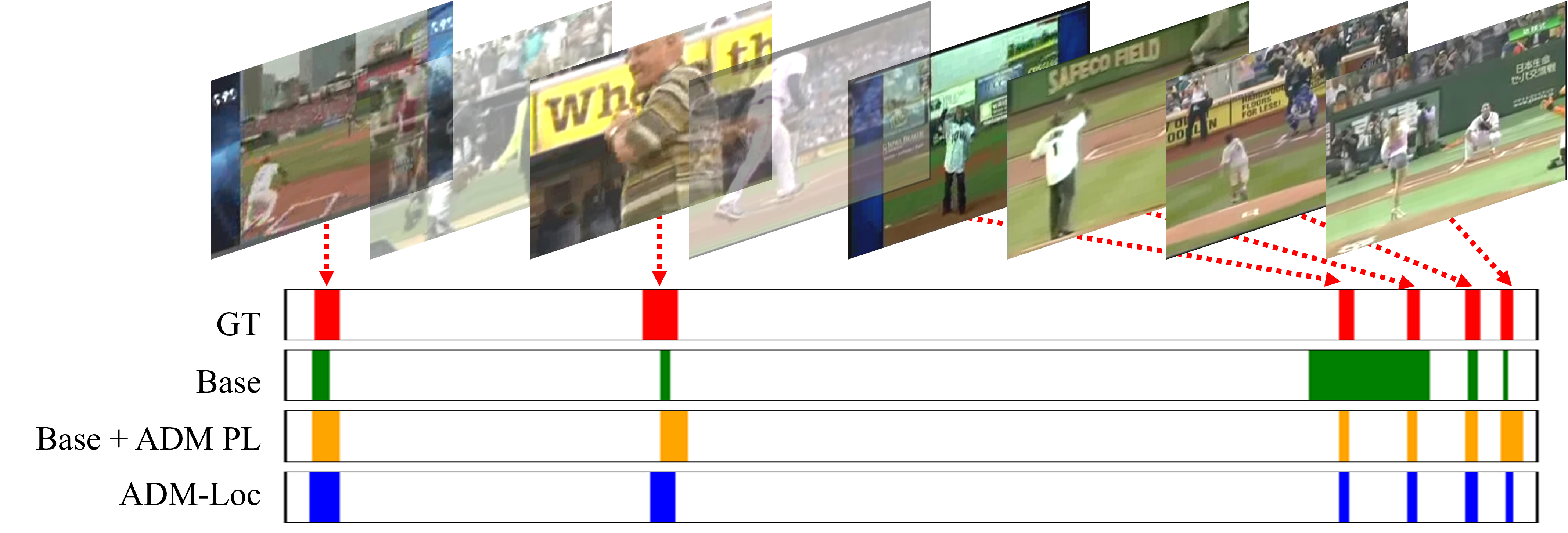}
        \caption{Action ``BaseballPitch".}
    \end{subfigure}
    \caption{Qualitative results on THUMOS'14. The ground-truth instances are highlighted in red. The detection results are displayed from: (1) the base model supervised with point-level annotations (green), (2) the base model supervised with our generated pseudo-labels (brown), (3) our ADM-Loc framework (blue). Transparent frames represent background frames.}
    \label{fig:vis_results}
\end{figure*}

Qualitative results depicted in Figure \ref{fig:vis_results} illustrate various types of errors, including over-completeness, incompleteness, and misalignment, generated by the base model. These issues have been addressed in ADM-Loc.

{
    \small
    \bibliographystyle{ieeenat_fullname}
    \bibliography{main}
}

\end{document}